\documentclass{article}
\pdfoutput=1
\PassOptionsToPackage{numbers, compress}{natbib}
\usepackage[preprint]{neurips_2024}
\usepackage[utf8]{inputenc} % allow utf-8 input
\usepackage[T1]{fontenc}  % use 8-bit T1 fonts
\usepackage{xcolor}     % colors
\usepackage{color}
\definecolor{ggreen}{HTML}{58a55c}
\usepackage[pagebackref=true,
   breaklinks=true,
   letterpaper=true,
   colorlinks = true,
   linkcolor = red,
   urlcolor = blue,
   citecolor = ggreen,
   anchorcolor = blue]{hyperref}
\usepackage{url}      % simple URL typesetting
\usepackage{booktabs}    % professional-quality tables
\usepackage{amsfonts}    % blackboard math symbols
\usepackage{nicefrac}    % compact symbols for 1/2, etc.
\usepackage{microtype}   % microtypography

\usepackage{algorithmic}
\usepackage{algorithm}

\usepackage{amsmath}
\usepackage{graphicx}
\usepackage{algorithm}
\usepackage{subcaption}
\usepackage{xcolor}
\usepackage{colortbl}

\usepackage{pifont}
\usepackage{tabu}
\usepackage{tabularx}
\definecolor{Gray}{gray}{0.9}
\definecolor{Cyan}{rgb}{0.88,1,1}
\usepackage{listings}
\usepackage{bm}
\usepackage{adjustbox}
\usepackage{ragged2e}
\usepackage{multirow}
\usepackage{amssymb}
\usepackage{wrapfig}
\usepackage{makecell}

\usepackage{soul}
\sethlcolor{Gray}

\numberwithin{equation}{section}

\newcolumntype{x}[1]{>{\centering\arraybackslash}p{#1pt}}
\newlength\savewidth

\newcommand{\PreserveBackslash}[1]{\let\temp=\\#1\let\\=\temp}
\newcolumntype{C}[1]{>{\PreserveBackslash\centering}p{#1}}
\newcolumntype{L}[1]{>{\PreserveBackslash\raggedright}p{#1}}

\usepackage{etoolbox}
\makeatletter
\AfterEndEnvironment{algorithm}{\let\@algcomment\relax}
\AtEndEnvironment{algorithm}{\kern2pt\hrule\relax\vskip3pt\@algcomment}
\let\@algcomment\relax
\newcommand\algcomment[1]{\def\@algcomment{\footnotesize#1}}
\renewcommand\fs@ruled{\def\@fs@cfont{\bfseries}\let\@fs@capt\floatc@ruled
 \def\@fs@pre{\hrule height.8pt depth0pt \kern2pt}%
 \def\@fs@post{}%
 \def\@fs@mid{\kern2pt\hrule\kern2pt}%
 \let\@fs@iftopcapt\iftrue}
\makeatother

\newcommand*\samethanks[1][\value{footnote}]{\footnotemark[#1]}

\title{H-vmunet: High-order Vision Mamba UNet for Medical Image Segmentation}

\author{
Renkai Wu$^{1}$
~~
Yinghao Liu$^{2}$ 
~~
Pengchen Liang$^{1}$\thanks{Corresponding authors.}
~~
Qing Chang$^{1}$\samethanks
\\
$^{1}$Shanghai University ~~ $^{2}$University of Shanghai for Science and Technology
}

\begin{document}

\maketitle

\begin{abstract}
In the field of medical image segmentation, variant models based on Convolutional Neural Networks (CNNs) and Visual Transformers (ViTs) as the base modules have been very widely developed and applied. However, CNNs are often limited in their ability to deal with long sequences of information, while the low sensitivity of ViTs to local feature information and the problem of secondary computational complexity limit their development. Recently, the emergence of state-space models (SSMs), especially 2D-selective-scan (SS2D), has had an impact on the longtime dominance of traditional CNNs and ViTs as the foundational modules of visual neural networks. In this paper, we extend the adaptability of SS2D by proposing a High-order Vision Mamba UNet (H-vmunet) for medical image segmentation. Among them, the proposed High-order 2D-selective-scan (H-SS2D) progressively reduces the introduction of redundant information during SS2D operations through higher-order interactions. In addition, the proposed Local-SS2D module improves the learning ability of local features of SS2D at each order of interaction. We conducted comparison and ablation experiments on three publicly available medical image datasets (ISIC2017, Spleen, and CVC-ClinicDB), and the results all demonstrate the strong competitiveness of H-vmunet in medical image segmentation tasks. The code is available from~\url{https://github.com/wurenkai/H-vmunet}.
\end{abstract}

\section{Introduction}
The rapid development of deep learning has led to the wide application of computer vision in medical image segmentation. Existing image segmentation mainly relies on two large base modules, convolution and Vision Transformers \cite{dosovitskiy2020image}. However, it is difficult for the convolution-based methods to acquire remote information \cite{rao2022hornet, rao2021global, wu2023automatic}, and even though convolution with a large convolution kernel \cite{liu2022convnet, ding2022scaling} is used for processing remote information, the results are still unsatisfactory. For the Transformers-based method, researchers \cite{dosovitskiy2020image} proposed to use Vision Transformers to extract image information. And Transformers for efficient data utilization (DeiT) was proposed in Touvron et al. \cite{touvron2021training}. However, Transformers-based methods usually need to occupy a large amount of memory.

Recently, state-space modeling (SSM) has gradually attracted a lot of attention. This is because SSMs are good at capturing remote dependencies and are capable of parallel training. The trained model parameters can be greatly reduced by parallel training. Nowadays, SSM methods are mainly from classical state space models \cite{kalman1960new}, which include Linear State Space Layer (LSSL) \cite{gu2021combining}, Structured State Space Sequence Model (S4) \cite{gu2021efficiently} and so on. They all have a very good result in capturing remote dependencies. In particular, researchers have proposed a Mamba \cite{gu2023mamba} algorithmic model by incorporating time-varying parameters into SSM. Mamba can achieve a better performance with lower parameters than traditional Transformers. In Mamba, the researchers demonstrated that Mamba will probably be a future alternative to Transformers in text modeling. Moreover, in Zhu et al. \cite{zhu2024vision}, researchers first proposed to introduce Mamba into image recognition by proposing Vision Mamba (Vim). Vim can save 86.8$\%$ of GPU memory when reasoning in images of 1248$\times$1248 size without attention. In particular, Vim computes recognition for high-resolution vision tasks better than traditional convolution and Transformers, thanks to Mamba's superior hardware-aware design \cite{zhu2024vision}.

In this paper, we extend the applicability of SSM. In 2D images, we specifically refer to the 2D-selective-scan (SS2D) \cite{liu2024vmamba} operation of SSM. SS2D is able to achieve a global receptive field at the cost of maintaining linear complexity and exhibits good feature extraction. However, the same attention to global receptive fields introduces more redundant information that is not relevant to the target features. Inspired by Rao et al. \cite{rao2022hornet}, we propose to use High-order SS2D (H-SS2D), which is a step-by-step phased approach to SS2D. This can minimize the introduction of redundant information while guaranteeing an excellent global receptive field for SS2D. In addition, we propose a High-order visual state space (H-VSS) module, of which the most critical module is H-SS2D.

Furthermore, the most classical framework for medical image segmentation, UNet \cite{ronneberger2015u}, has a very good information integration. UNet is composed of an encoder, a skip-connection path and a decoder. Traditionally, UNet uses full convolution and downsampling to extract feature information, and uses upsampling and convolution to recover the image size. In particular, the skip-connection path of UNet can effectively complement the lost detail information due to downsampling, which is crucial for medical image segmentation. Since much of the detail information in medical images is very crucial, the skip-connection path plays an important role in improving the performance of medical image segmentation. Currently, researchers have developed many variants of the UNet framework, such as U-Net v2 \cite{peng2023u}, META-Unet \cite{wu2023meta}, Att UNet \cite{oktay2018attention}, MALUNet \cite{ruan2022malunet} and so on. These all demonstrate the effectiveness of the UNet framework for medical image segmentation. In particular, for High-order spatial interaction methods, researchers proposed MHorUNet \cite{wu2024mhorunet}, HSH-UNet \cite{wu2024hsh}, MHA-UNet \cite{wu2023only}, and so on. For the VMamba method, in Ruan et al. \cite{ruan2024vm}, researchers firstly proposed a pure Vision Mamba based UNet model (VM-UNet) for medical image segmentation. Inspired by this study, we adopt the H-VSS proposed in this study to construct the U-shaped model to form the High-order Vision Mamba UNet (H-vmunet). To the best of our knowledge, H-vmunet is the first to introduce VMamba to higher order for processing feature information. And higher-order has cleaner feature information and less redundant information.

Our contributions can be summarized as follows:

\begin{itemize}
\item Based on the state-space model (SSM), we propose a High-order 2D-selective-scan (H-SS2D). The High-order operation ensures the excellent global sensory field under SS2D while minimizing the introduction of redundant information.

\item Based on the proposed H-SS2D, we construct the High-order visual state space (H-VSS) module of H-SS2D.

\item Based on the proposed H-VSS module, we constructed the High-order Vision Mamba UNet (H-vmunet) for medical image segmentation. To the best of our knowledge, the H-vmunet module is the first framework that introduces Vision Mamba to High-order operations.

\item Our proposed H-vmunet reduces the number of parameters by 67.28$\%$ over the traditional Vision Mamba UNet model (VM-UNet), and the performance is significantly improved in all three publicly available medical image datasets.

\end{itemize}

\section{Related work}
\subsection{Image segmentation}
Image segmentation, as an important field branch of computer vision, plays an important role in various industries. In the past when computer hardware was inadequate, traditional image segmentation was a mainstream method. Traditional methods generally refer to the use of simple mathematical image processing techniques and mathematical methods to differentiate between places where the thresholds are significantly different \cite{maurya2023review}, and to form a clear differentiation in each region \cite{liu2021review}. Traditional methods tend to have low hardware requirements, but for the need to refine the region segmentation and in the complex conditions of the segmentation often meet the demand. With the breakthrough of hardware, the time for computers to realize complex calculations is significantly reduced. In addition, in terms of algorithmic models, the emergence of the full convolutional model (FCN) \cite{long2015fully} broke the long-standing dominance of traditional image segmentation methods. FCN is also the first network model that introduces deep learning methods into the field of image segmentation. Afterwards, with the emergence of FCN, more and more researchers of traditional image segmentation methods turned to the research of deep learning image segmentation methods. Among them, the emergence of U-Net \cite{ronneberger2015u} model has led to the rapid development of deep learning image segmentation methods in the field of medical images. This is due to the fact that the skip-connection path in U-Net well preserves and fuses high-level and low-level features, which is crucial for medical image segmentation that requires refinement.

\subsection{Medical image segmentation}
Medical image segmentation requires more lesion feature learning compared to image segmentation in natural scenes. In particular, solving the complex scale variation problem is an important issue in medical image segmentation \cite{xiao2023transformers}. The U-Net framework has the ability to combine low-level and high-level features, which makes it occupy a major position in the field of medical image segmentation.

After the proposal of U-Net, more and more medical image segmentation researchers have proposed many variant models of the U-Net framework. In Peng et al. \cite{peng2023u}, researchers proposed a novel skip-connection approach to feature refinement by mixing multiple high-level and low-level features. This allows for a better combination of feature information at multiple scales for learning. In Oktay et al.\cite{oktay2018attention}, the researchers added to the U-Net framework through the attention mechanism, which makes the traditional U-Net framework model increase the weight of the focused object and suppress irrelevant information. ATTENTION SWIN U-Net \cite{aghdam2023attention} operates in a cascade of the attention mechanism and is based on Swin U-Net \cite{cao2022swin} for tuning optimization. TransNorm \cite{azad2022transnorm} utilized both Convolution and Transformers for constructing a U-Net-like framework, with the jump-connectivity portion using an adaptive calibration of features with an attention mechanism. In Ruan et al. \cite{ruan2022malunet}, researchers proposed an ultra-lightweight U-Net model, specifically, using multilevel and multiscale fusion mechanisms to minimize the need for the number of input channels. MSNet \cite{zhao2021automatic} proposed a novel skip-connection path using an unusual subtraction operation for fusing feature information at different scales, which avoids the redundancy introduced by excessive summing or splicing. introduced redundant information. SCR-Net \cite{wu2021precise} proposes a semantic calibration module and a semantic refinement module based on modules under convolution to improve the learning ability of features. In Wu et al. \cite{wu2023meta}, researchers used two Transformers with different resolutions to learn features at different scales separately, and the results showed that this approach can effectively improve the model's fine-grained feature learning ability and generalization ability. C$^2$SDG \cite{hu2023devil} proposed a contrast-enhanced feature learning method, which performed low-level feature and style enhancement and used for subsequent contrast training. MHorUNet \cite{wu2024mhorunet} introduced the higher-order spatial interaction mechanism \cite{rao2022hornet} into the framework of U-Net for the first time and achieved excellent segmentation performance in skin lesions.

Recently, the emergence of Vision Mamba (VMamba) has led to its presence in the field of medical image segmentation. VM-UNet \cite{ruan2024vm} is the first model that introduces VMamba into the U-Net framework for pure VMamba and is used for medical image segmentation. The effectiveness of VM-UNet is demonstrated based on experimental results in ISIC2017 \cite{codella2018skin}, ISIC2018 \cite{codella2019skin} and Synapse \cite{landman2015miccai} datasets. However, VM-UNet still has some problems compared to other current state-of-the-art medical segmentation models, such as excessive memory usage and lack of performance.

In this paper, we introduce VMamba to higher-order extraction of features, which can enhance the extraction of target features and reduce feature fusion with redundant information under the condition of maximizing the utilization of the global receptive field of SS2D. In addition, we propose High-order Vision Mamba UNet (H-vmunet) for medical image segmentation in conjunction with the U-Net framework. In the next section, we will elaborate each module in detail.

\begin{figure}[!t]
\centering
\includegraphics[width=\linewidth]{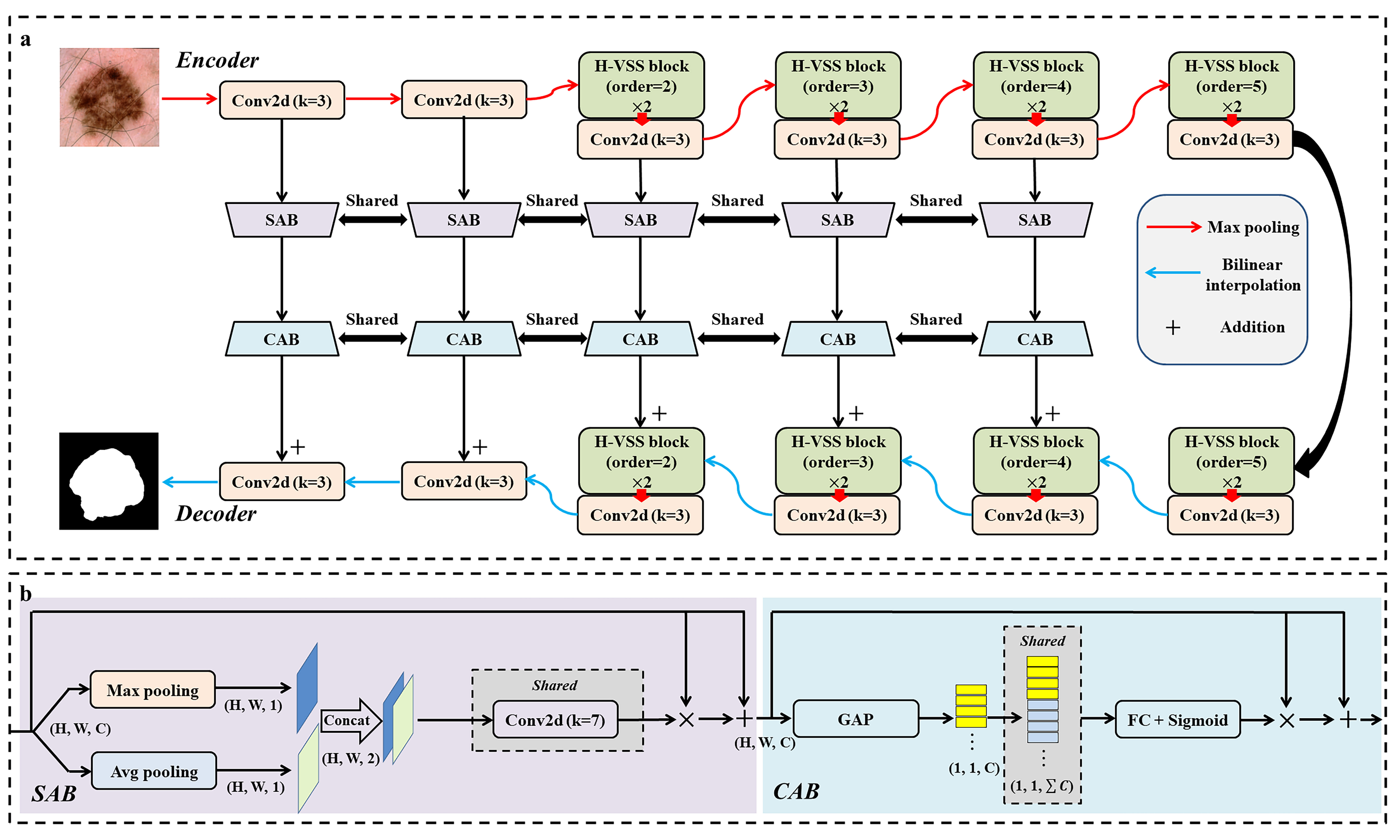}
\caption{(a) The proposed High-order Vision Mamba UNet (H-vmunet) model architecture. (b) Multi-level and multi-scale information fusion module architecture.}
\label{fig01}
\end{figure}

\begin{figure}[!t]
\centering
\includegraphics[width=\linewidth]{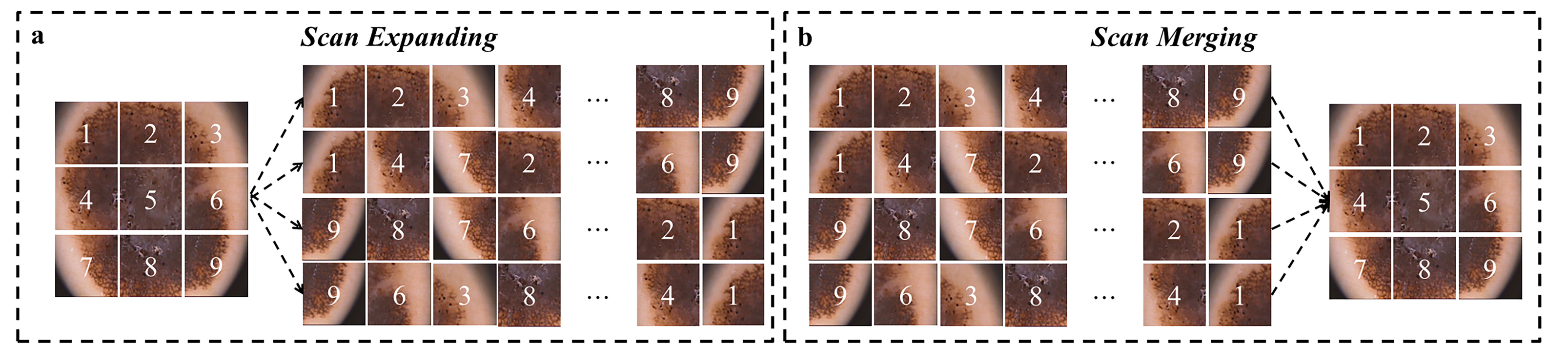}
\caption{Image description for 2D-selective-scan.}
\label{fig02}
\end{figure}

\section{Method}
\subsection{Architecture overview}
Our proposed High-order Vision Mamba UNet (H-vmunet) model is shown in Figure \ref{fig01}. The proposed H-vmunet has a total of 6 layers. The proposed H-vmunet has a total of 6-layer structure with U-shaped architecture, which mainly consists of encoder, decoder, and skip-connect parts. The number of channels in the 6-layer structure is set to [8,16,32,64,128,256]. Layers 1 and 2 use a standard convolution with a convolution kernel of 3, respectively. Layers 3 to 6 are the most central part of the H-vmunet, respectively, and each layer is set up with a High-order visual state space (H-VSS) module and a standard convolution with a convolution kernel of three. In particular, the H-VSS of order 2,3,4,5 is used from layer 3 to 6, respectively. Specifically, the H-VSS serves as the core of our proposed model, which will be detailed in the next subsections. The H-vmunet skip-connection path utilizes the Channel Attention Bridge (CAB) module and Spatial Attention Bridge (SAB) module for multilevel and multiscale information fusion \cite{ruan2022malunet}. The operation of multilevel and multiscale information fusion improves the convergence speed of the model and increases the sensitivity to lesions at different scales.

\subsection{2D-selective-scan}
In order to introduce High-order 2D-selective-scan, we first describe the 2D-selective-scan proposed by the previous authors \cite{liu2024vmamba}. As shown in Figure \ref{fig02}, a schematic diagram of the main operations of SS2D is shown, which includes the scan expansion operation, the S6 block operation and the scan merge operation. As shown in Figure \ref{fig02}(a), the scan expansion operation is shown, in which the expansion sequence is performed in four different scan order directions from top-left to bottom-right, bottom-right to top-left, top-right to bottom-left, and bottom-left to top-right, respectively. Then, feature extraction is performed in each direction by S6 block \cite{gu2023mamba}. Where the specific feature extraction method of the S6 block can be derived from the Algorithm \ref{alg:s6}. Finally, the sequences in four different directions are merged by performing a scan merge operation as shown in Figure \ref{fig02}(b) and returned to the initial image size. For a more detailed description of SS2D, the reader is advised to refer to \cite{gu2023mamba}.

\subsection{High-order visual state space}
\begin{figure}[!t]
\centering
\includegraphics[width=\linewidth]{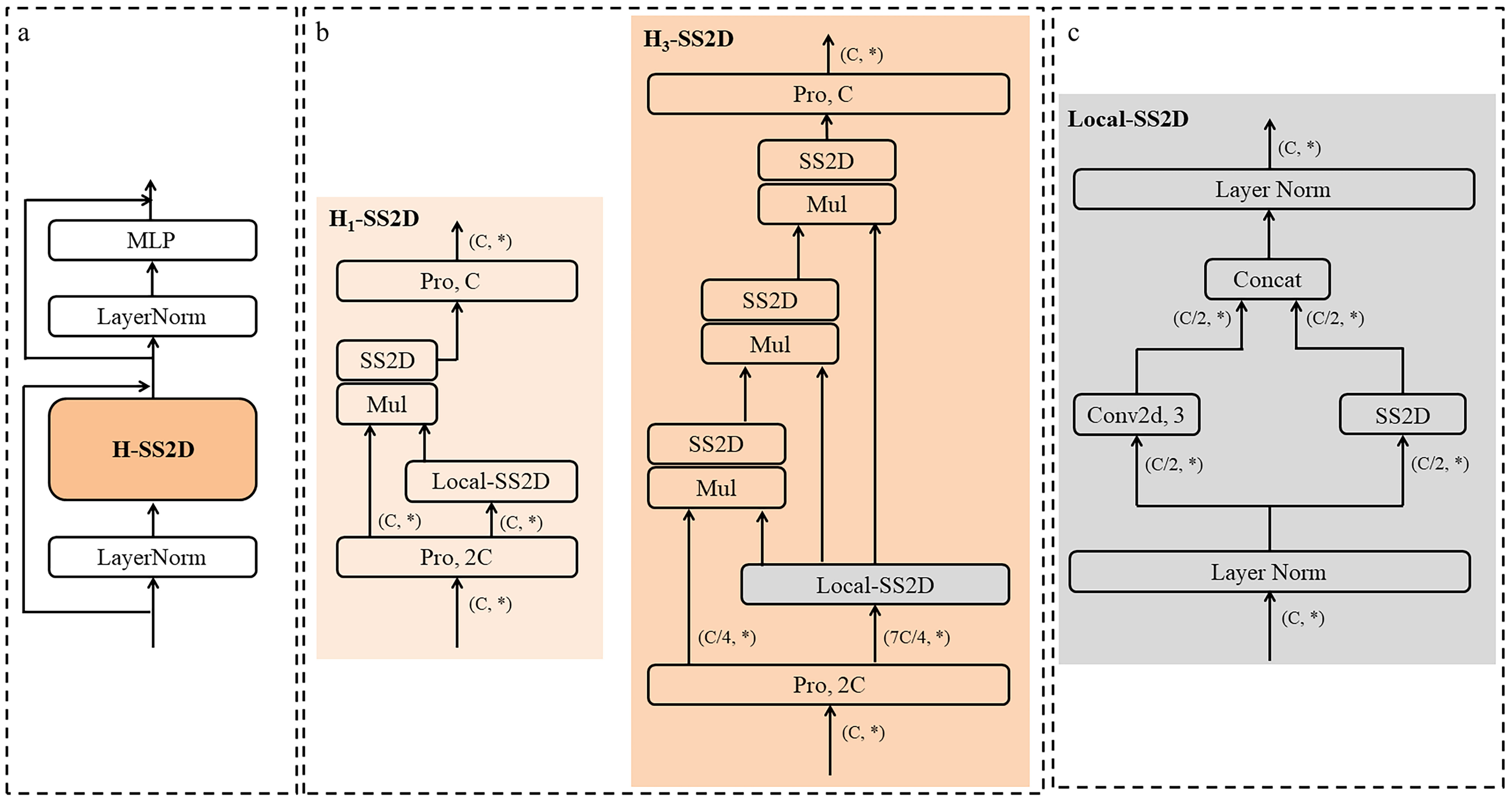}
\caption{(a) The proposed High-order visual state space (H-VSS) module architecture. (b) Overview of 1-order and 3-order 2D-selective-scan (H$_1$-SS2D and H$_3$-SS2D). (c) Overview of the proposed Local-SS2D module.}
\label{fig03}
\end{figure}

The proposed High-order visual state space (H-VSS) module is shown in Figure \ref{fig03}(a). We adopt a similar structure as Transformers, where the self-attention layer is replaced by the proposed High-order 2D-selective-scan (H-SS2D). The specific H-VSS expression can be expressed by the following equations: 
\begin{equation}
y=H S[L N(x)]+x
\end{equation}
\begin{equation}
Out=M L P[L N(y)]+y
\end{equation}
\noindent
where $LN$ is the LayerNorm layer, $HS$ is the H-SS2D operation, and $MLP$ is the multilayer perceptron. $HS$ is the key core component of which includes the visual state space for spatial interactions of arbitrary order, which is detailed in the next subsections.

\begin{algorithm}[!t]
\caption{Pseudo-code for S6 block \cite{gu2023mamba,ruan2024vm}}
\label{alg:s6}
\renewcommand{\algorithmicensure}{\textbf{Output:}}
\begin{algorithmic}[1]
\renewcommand{\algorithmicrequire}{\textbf{Input:}}
\REQUIRE $x$, the feature with shape [B, L, D] (batch size, token length, dimension) %%input
\renewcommand{\algorithmicrequire}{\textbf{Params:}}
\REQUIRE $\mathbf{A}$, the nn.Parameter; $\mathbf{B}$, the nn.Parameter; $\mathbf{C}$, the nn.Parameter; $\mathbf{D}$, the nn.Parameter
\renewcommand{\algorithmicrequire}{\textbf{Operator:}}
        \REQUIRE Linear(.), the linear projection layer
\ENSURE $y$, the feature with shape [B, L, D] %%output
        \STATE  $\mathbf{\Delta}, \mathbf{B}, \mathbf{C}$ = Linear($x$), Linear($x$), Linear($x$)
        \STATE  $\overline{\mathbf{A}} = \textup{exp}(\mathbf{\Delta} \mathbf{A})$
        \STATE  $\overline{\mathbf{B}} = (\mathbf{\Delta} \mathbf{A})^{-1}(\textup{exp}(\mathbf{\Delta} \mathbf{A}) - \mathbf{I})\cdot\mathbf{\Delta} \mathbf{B}$
        \STATE  $h_t = \overline{\mathbf{A}}h_{t-1} + \overline{\mathbf{B}}x_t$
        \STATE  $y_t = \mathbf{C}h_t + \mathbf{D}x_t$
        \STATE  $y = [y_1, y_2, \cdots, y_t, \cdots, y_L]$
        \RETURN $y$
\end{algorithmic}
\end{algorithm}

\subsubsection{1-order 2D-selective-scan}
For the sake of clarity, we describe the 1-order 2D-selective-scan (H$_1$-SS2D) operation in detail. The H1-SS2D shown in Figure \ref{fig03}(b) is the component structure of the 1-order 2D-selective-scan. The H$_1$-SS2D contains the linear projection layer, the Local-SS2D module, the SS2D operation, the Hadamard Product operation, etc. Among them, Local-SS2D module is the one we propose to enhance the local feature extraction capability of SS2D.

The Local-SS2D module is shown in Figure \ref{fig03}(c), which mainly consists of two pathways, including LayerNorm layer, standard convolution with convolution kernel of 3, SS2D operation and Concat operation. After the Local-SS2D module features will be half of the channel for the convolution operation, half of the channel for the SS2D operation, and finally the Concat operation for output. Combining the excellent local feature learning ability of convolution and the excellent learning ability of SS2D in the global sensory field, it can enhance the network's comprehensive feature learning ability. The specific Local-SS2D module can be expressed by the following equations:
\begin{equation}
x_{1}, x_{2}=S p[L N(x)]
\end{equation}
\begin{equation}
\left.y=Cat\left\{Conv\left(x_{1}\right), SS2D(x_{2}\right)\right\}
\end{equation}
\begin{equation}
Out =L N(y)
\end{equation}
\noindent
where $LN$ is the LayerNorm layer, $Sp$ is the Split operation (forming two features with half a channel each), and $Cat$ is the Concat operation.

In H$_1$-SS2D, features with 2 times the number of channels will be formed after the linear projection layer. After that, half the number of channels will be fed into the Local-SS2D module. The remaining half channel number of features will be Hadamard product with the output features passing through the Local-SS2D module, and SS2D operation will be performed again, and finally output through the linear projection layer. The specific operation can be expressed by the following equations:
\begin{equation}
\left[X^{H W \times C}, Y^{H W \times C}\right]=Pro(x) \in \mathbb{R}^{H W \times 2C}
\end{equation}
\begin{equation}
y=S S 2 D\left[X \odot LSD\left(Y\right)\right]
\end{equation}
\begin{equation}
Out=Pro(y) \in \mathbb{R}^{H W \times C}
\end{equation}
\noindent
where $Pro$ is the linear projection layer, $LSD$ is the Local-SS2D module, and $\odot$ is the Hadamard product operation. Through the above series of operations, which is the 1-order 2D-selective-scan operation, the local feature extraction ability of the network can be improved by the Local-SS2D module. However, the 1-order operation is weak in terms of H-SS2D's ability to reduce the introduction of redundant feature information. High-order 2D-selective-scan will be introduced next.

\subsubsection{High-order 2D-selective-scan}
By detailing the 1-order 2D-selective-scan operation, we can similarly generalize it to the $n$-order (H$_n$-SS2D), such as the H$_n$-SS2D of Figure \ref{fig03}(b). where the input features (with a channel number of C), after passing through the linear projection layer, will form a feature $X_0$ and a set of features $Y$ ($Y=[Y_0,Y_1,...,Y_n ]$), which have a total number of channels of 2C. For each order of features the number of channels can be calculated from $C_{k}=\frac{C}{2^{n-k-1}}, 0 \leq k \leq n-1$. The specific operation can be expressed by the following equations:
\begin{equation}
\left[X_{0}^{H W \times C_{0}}, Y_{0}^{H W \times C_{0}}, \cdots, Y_{n-1}^{H W \times C_{n-1}}\right]=Pro(x) \in \mathbb{R}^{H W \times 2C}
\end{equation}
\begin{equation}
X_{k+1}=S S 2 D\left[X_k \odot LSD\left(Y_k\right)\right]
\end{equation}
\begin{equation}
Out=Pro(X_n) \in \mathbb{R}^{H W \times C}
\end{equation}
\noindent
where $Pro$ is the linear projection layer, $LSD$ is the Local-SS2D module, and $\odot$ is the Hadamard product operation. As seen in Equation 3.10, for each order of interaction, the feature $X_k$ of the corresponding number of channels will be subjected to a Hadamard product operation with the feature that passes through the output of the Local-SS2D module, followed by the SS2D operation. A portion of the number of channels is used in each order to perform the above operation until the end of the nth order. Hadamard product with the features that have been output by the Local-SS2D module will increase the weights of the local features, and another SS2D will be performed to re-attend to the global sensory field, which will reduce the introduction of redundant information and increase the weights of the features on the target. In addition, such an operation will be performed $n$ times, gradually reducing the weight of redundant information.

\subsubsection{Multi-level and multi-scale information fusion module}
The multilevel and multiscale information fusion module mainly consists of the Channel Attention Bridge (CAB) module and Spatial Attention Bridge (SAB) module proposed by Ruan et al. \cite{ruan2022malunet} Specifically, a schematic diagram of the CAB module and SAB module is shown in Figure \ref{fig01}(b). The SAB module uses maximum pooling and average pooling operations to stitch together to form the feature maps of the two channels, and generates the spatial attention maps using extended convolution with shared weights and Sigmoid function. The residuals are connected through the spatial attention map to guide the attention of the features. The CAB module concatenates feature information of different stages and scales on different channel axes and generates the corresponding channel attention maps, which in turn guide the residual fusion of subsequent features. The fused features after the SAB and CAB modules can improve the richness of feature information scales during decoder fusion.

\section{Experiment}
\subsection{Datasets}
To validate the effectiveness of the proposed H-vmunet, we conducted experiments on three publicly available medical image datasets, respectively. Among them, the datasets include the skin lesion dataset, the Spleen dataset and the polyp dataset.

The skin lesion dataset is the ISIC2017 \cite{codella2018skin} dataset published by the International Skin Imaging Collaboration (ISIC). Specifically, we obtained 2000 dermoscopic images along with the corresponding segmentation mask labels from the ISIC2017 dataset. The initial size of the images is 576$\times$767, and the preprocessing uniform size is 256$\times$256. where 1250 images are arbitrarily divided for training, 150 images are used as validation, and the remaining 600 images are used as testing.

The Spleen dataset was released using Antonelli et al. \cite{antonelli2022medical} to release the Spleen dataset, sourced from Memorial Sloan Kettering Cancer Center, which includes, 41 cases for the training set and 20 cases for the test set. Since the labels of the test set were not officially provided, we only used 40 cases to randomly divide the training, validation and test sets of our experiments. The training set includes 32 cases (841 images), the validation set has 3 cases (48 images), and the remaining 6 cases are used as the test set (160 images). The image size is uniformly 320$\times$320.

The polyp dataset utilizes the CVC-ClinicDB \cite{bernal2015wm} provided by the MICCAI 2015 automated polyp detection subtask. In total, the dataset contains 612 still images extracted from colonoscopy videos and the corresponding segmentation mask labels. The initial size of the images is 576$\times$768 and the preprocessed uniform size is 256$\times$256. We arbitrarily divide the training, validation and test sets, where the training set contains 429 images, the validation set has 62 images, and the remaining 121 images are used as the test set.

\subsection{Implementation details}
All experiments were implemented based on Python 3.8 and Pytorch 1.13.0, and were run on a single NVIDIA V100 GPU with 32 GB of memory. To increase the diversity of the data, we stayed consistent with \cite{wu2024mhorunet} and used horizontal and vertical flips, as well as random rotation operations. For skin lesion and polyp segmentation, the image size is uniformly 256$\times$256. For Spleen segmentation, the image size is uniformly 320$\times$320. The loss function is adopted from BceDice loss function \cite{wu2024mhorunet, ruan2022malunet, wu2024hsh, wu2023only}. The training epoch is set to 250 and the batch size is 8. AdamW \cite{loshchilov2017decoupled} is used as the optimizer with an initial learning rate of 0.001, and cosine annealing is used to learn the scheduler, with the minimum value of the learning rate set to 0.00001.

\subsection{Evaluation metrics}
Medical image segmentation generally uses four evaluation metrics to judge the model segmentation performance. These include dice similarity coefficient (DSC), sensitivity (SE), specificity (SP), and accuracy (ACC). DSC is mainly used to rate the similarity of predicted segmentation values to the ground truth. SE is mainly used to measure the percentage of true positives among true positives and false negatives. SP is mainly used to measure the percentage of true negatives among true negatives and false positives. ACC is mainly used to judging for the percentage of correct categorization. The expression can be expressed by the following equations:
\begin{equation}
\mathrm{D S C}=\frac{2\mathrm{TP}}{2\mathrm{TP}+\mathrm{FP}+\mathrm{FN}}
\end{equation}
\begin{equation}
\mathrm{A C C}=\frac{\mathrm{TP}+\mathrm{TN}}{\mathrm{TP}+\mathrm{TN}+\mathrm{FP}+\mathrm{FN}}
\end{equation}
\begin{equation}
\mathrm{Sensitivity} =\frac{\mathrm{TP}}{\mathrm{TP}+\mathrm{FN}}
\end{equation}
\begin{equation}
\mathrm{Specificity} =\frac{\mathrm{TN}}{\mathrm{TN}+\mathrm{FP}}
\end{equation}
\noindent
where TP denotes true positive, TN denotes true negative, FP denotes false positive and FN denotes false negative.

\begin{figure}[!t]
\centering
\includegraphics[width=0.9\linewidth]{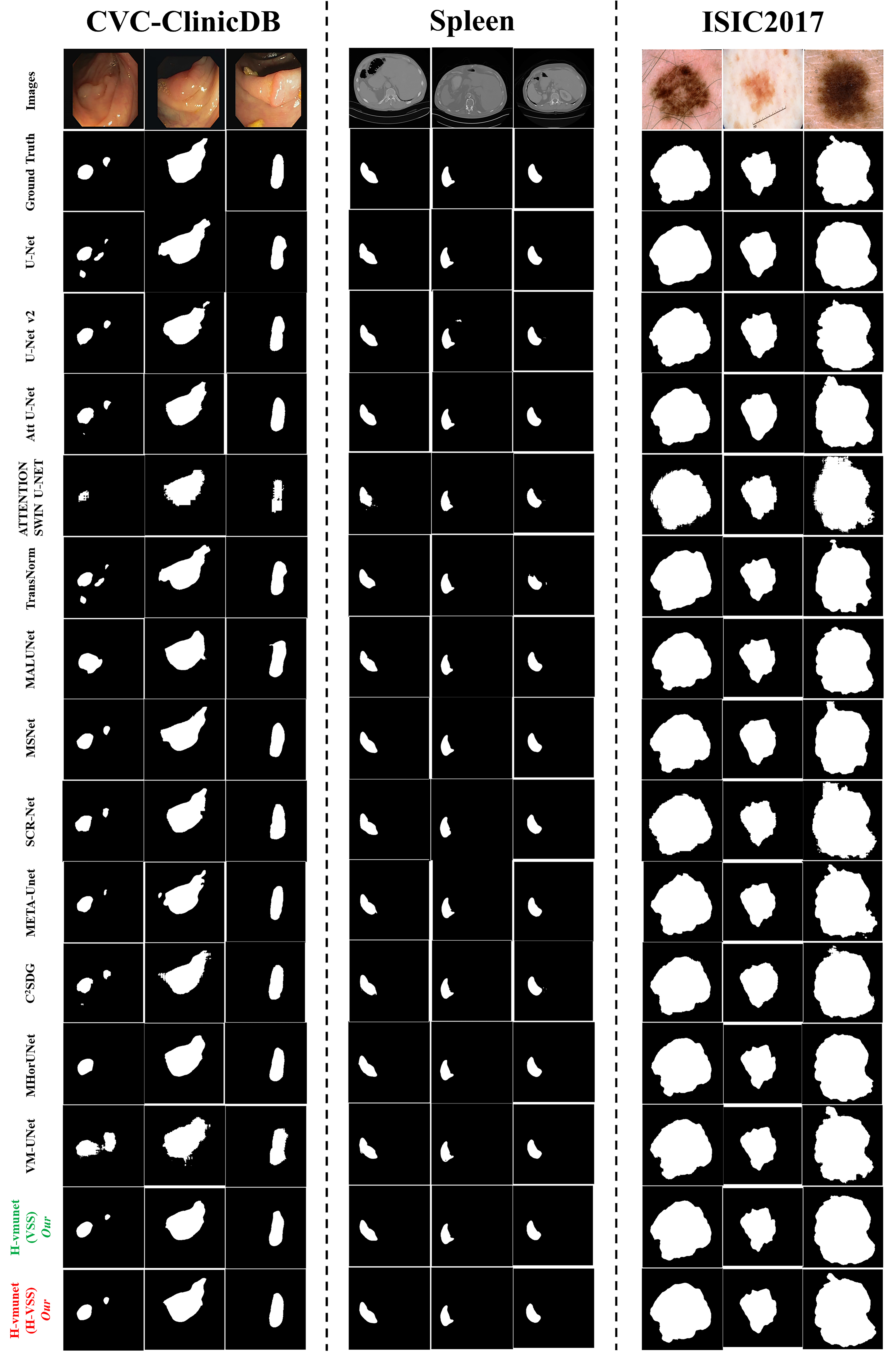}
\caption{Visualization of segmentation graphs for comparison experiments.}
\label{fig04}
\end{figure}

\subsection{Comparison results}
In order to demonstrate the superior performance of our proposed H-vmunet model, we conducted comparison experiments with several of the most popular and advanced medical image segmentation models. Specifically, they include U-Net \cite{ronneberger2015u}, U-Net v2 \cite{peng2023u}, Att U-Net \cite{oktay2018attention}, ATTENTION SWIN U-Net \cite{aghdam2023attention}, TransNorm \cite{azad2022transnorm}, MALUNet \cite{ruan2022malunet}, MSNet \cite{zhao2021automatic}, SCR-Net \cite{wu2021precise}, META-Unet \cite{wu2023meta}, C$^2$SDG \cite{hu2023devil}, MHorUNet \cite{wu2024mhorunet} and VM-UNet \cite{ruan2024vm}.

\begin{table}[!t]
\centering
\caption{Results of comparison experiments on three publicly available medical image datasets.}
\resizebox{\linewidth}{!}{
\begin{tabular}{ccccccccccccc}
\hline
\multirow{2}{*}{\textbf{Methods}} & \multicolumn{4}{c}{\textbf{ISIC2017}}                                 & \multicolumn{4}{c}{\textbf{Spleen}}                                   & \multicolumn{4}{c}{\textbf{CVC-ClinicDB}}                             \\ \cline{2-13} 
                                  & \textbf{DSC}    & \textbf{SE}     & \textbf{SP}     & \textbf{ACC}    & \textbf{DSC}    & \textbf{SE}     & \textbf{SP}     & \textbf{ACC}    & \textbf{DSC}    & \textbf{SE}     & \textbf{SP}     & \textbf{ACC}    \\ \hline
U-Net \cite{ronneberger2015u}                             & 0.8159          & 0.8172          & 0.9680          & 0.9164          & 0.9441          & 0.9604          & 0.9989          & 0.9983          & 0.9039          & \textbf{0.8926} & 0.9914          & 0.9821          \\
U-Net v2 \cite{peng2023u}                          & 0.9149          & 0.9052          & 0.9821          & 0.9670          & 0.9383          & 0.9244          & 0.9990          & 0.9982          & 0.9055          & 0.8861          & 0.9930          & 0.9825          \\
Att U-Net \cite{oktay2018attention}                         & 0.8082          & 0.7998          & 0.9776          & 0.9145          & 0.9321          & 0.9350          & 0.9989          & 0.9979          & 0.8697          & 0.8474          & 0.9894          & 0.9760          \\
ATTENTION SWIN U-Net \cite{aghdam2023attention}              & 0.8859          & 0.8492          & 0.9847          & 0.9591          & 0.7829          & 0.6662          & \textbf{0.9994} & 0.9945          & 0.7526          & 0.6503          & \textbf{0.9943} & 0.9631          \\
TransNorm \cite{azad2022transnorm}                         & 0.8933          & 0.8535          & 0.9859 & 0.9582          & 0.8618          & 0.7938          & 0.9992          & 0.9962          & 0.8845          & 0.8634          & 0.9918          & 0.9801          \\
MALUNet \cite{ruan2022malunet}                           & 0.8896          & 0.8824          & 0.9762          & 0.9583          & 0.9310          & 0.9305          & 0.9989          & 0.9979          & 0.8562          & 0.8475          & 0.9862          & 0.9731          \\
MSNet \cite{zhao2021automatic}                             & 0.9067          & 0.8771          & \textbf{0.9860}          & 0.9647          & 0.9521          & 0.9425          & \textbf{0.9994} & 0.9986          & 0.9050          & 0.8720          & 0.9932          & 0.9832          \\
SCR-Net \cite{wu2021precise}                           & 0.8898          & 0.8497          & 0.9853          & 0.9588          & 0.9181          & 0.9122          & 0.9988          & 0.9976          & 0.8951          & 0.8701          & 0.9922          & 0.9807          \\
META-Unet \cite{wu2023meta}                         & 0.9068          & 0.8801          & 0.9836          & 0.9639          & 0.9233          & 0.8921          & 0.9993          & 0.9978          & 0.8975          & 0.8768          & 0.9919          & 0.9811          \\
C$^2$SDG \cite{hu2023devil}                             & 0.8938          & 0.8859          & 0.9765          & 0.9588          & 0.9354          & 0.9263          & 0.9991          & 0.9981          & 0.8967          & 0.8724          & 0.9923          & 0.9810          \\
MHorUNet \cite{wu2024mhorunet}                          & 0.9132          & 0.8974          & 0.9834          & 0.9666          & 0.9424          & 0.9508          & 0.9989          & 0.9982          & 0.8930          & 0.8803          & 0.9904          & 0.9801          \\
VM-UNet \cite{ruan2024vm}                           & 0.9070          & 0.8837          & 0.9842          & 0.9645          & 0.9418          & 0.9429          & 0.9991          & 0.9982          & 0.8524          & 0.8370          & 0.9867          & 0.9726          \\
\textbf{H-vmunet (VSS) (Our)}           & 0.9068          & 0.8897          & 0.9823          & 0.9642          & 0.9403          & 0.9330          & 0.9992          & 0.9982          & 0.8984          & 0.8768          & 0.9921          & 0.9813          \\
\textbf{H-vmunet (H-VSS) (Our)}         & \textbf{0.9172} & \textbf{0.9056} & 0.9831          & \textbf{0.9680} & \textbf{0.9571} & \textbf{0.9642} & 0.9992          & \textbf{0.9987} & \textbf{0.9087} & 0.8803          & 0.9940          & \textbf{0.9833} \\ \hline
\label{tab1}
\end{tabular}}
\end{table}

\begin{figure}[!t]
\centering
\includegraphics[width=\linewidth]{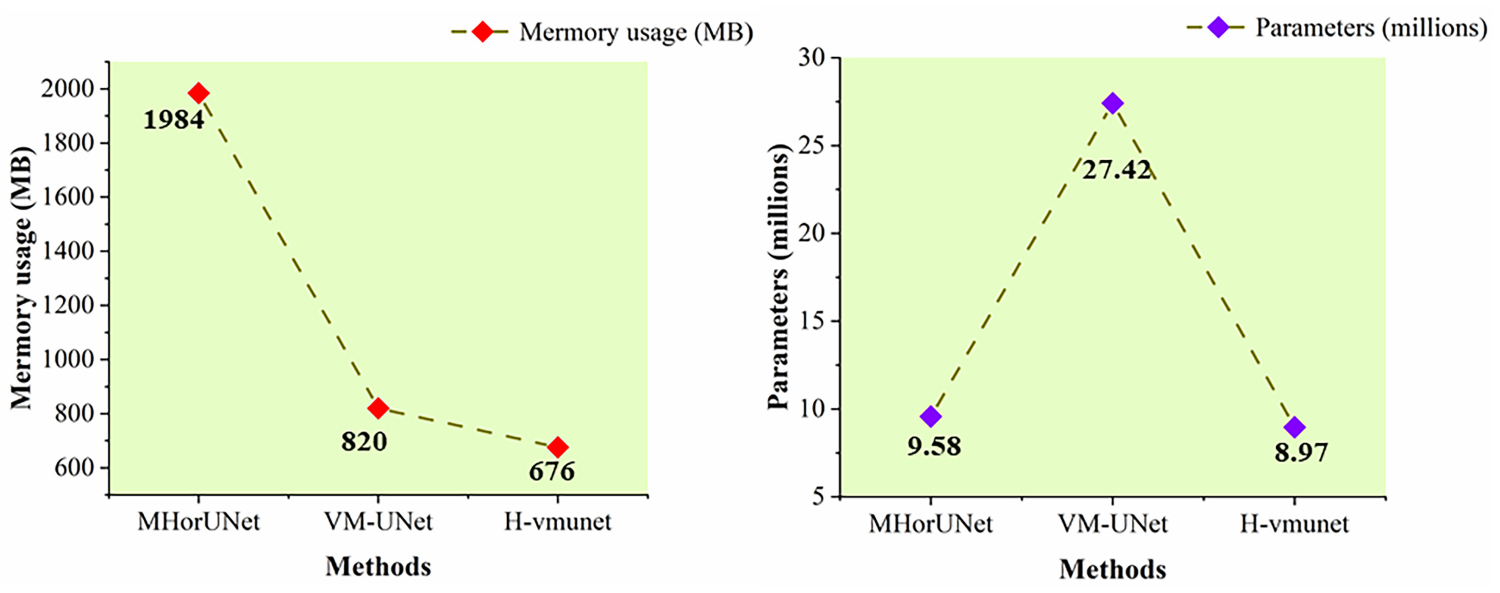}
\caption{Comparison of the parameters and memory usage of the proposed H-vmunet with the traditional High-order spatial interaction UNet (MHorUNet) model and the pure Vision Mamba UNet (VM-UNet) model.}
\label{fig05}
\end{figure}

As shown in Table \ref{tab1} are the experimental results on three publicly available medical image datasets. In particular, our proposed model H-vmunet uses two forms for comparison. H-vmunet (H-VSS) is our best performing model, while H-vmunet (VSS) is a direct replacement of SS2D in the traditional visual state space (VSS) \cite{liu2024vmamba} module using the proposed H-SS2D. Adding this comparison can again demonstrate that our proposed H-VSS has more efficient results than VSS access to H-SS2D. As can be seen from the table, the performance of the proposed H-vmunet is improved over the traditional higher-order spatial interaction UNet model (MHorUNet) in all three public datasets. From the table, it can be concluded that the performance of the proposed H-vmunet is improved over the traditional higher-order spatial interaction UNet model (MHorUNet) in all the three public datasets. This suggests that the proposed introduction of SS2D into higher order can further improve the effectiveness of higher-order spatial interaction. In addition, comparing with the traditional pure Vision Mamba UNet model (VM-UNet), we propose to introduce Vision Mamba into higher order to form the H-vmunet model, while it shows more excellent performance in all three publicly available medical image datasets. This confirms that Vision Mamba is better able to exhibit its superior global receptive field at higher orders, thanks to the fact that higher-order spatial interactions can gradually reduce the introduction of redundant information and improve the learning of local feature information. The visualized segmentation comparison result graph is shown in Figure \ref{fig04}. By visualizing the result graph, we can see more clearly that the proposed H-vmunet has the ability to be more sensitive to lesion information. While the traditional Vision Mamba UNet model (VM-UNet) still has deficiencies in learning the features of the lesions, and the contours are still blurred even though all of them can correctly recognize the location of the lesions.

As shown in Figure \ref{fig05}, it is the comparison of the parameters of the proposed H-vmunet with the traditional higher-order spatial interaction UNet model (MHorUNet) and the UNet model of pure Vision Mamba (VM-UNet). From the table, it can be concluded that H-vmunet reduces the number of parameters by 67.28$\%$ compared to VM-UNet and H-vmunet reduces the number of parameters by 6.37$\%$ compared to MHorUNet. 

\begin{figure}[!t]
\centering
\includegraphics[width=\linewidth]{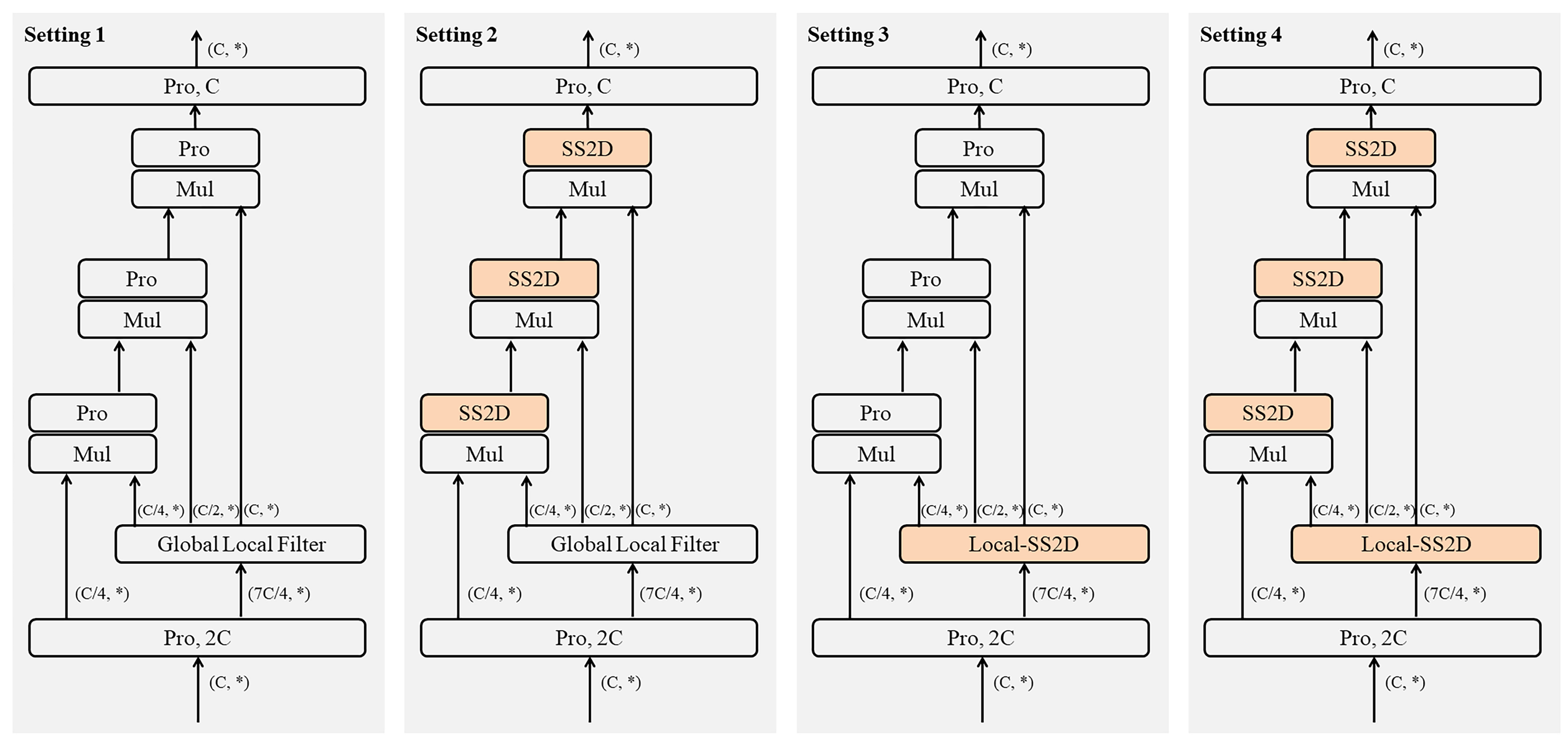}
\caption{Ablation experiments were conducted using four different settings to validate the effectiveness of employing SS2D manipulation in High-order spatial interaction mechanisms.}
\label{fig06}
\end{figure}

\subsection{Ablation experiment}
\begin{figure}[!t]
\centering
\includegraphics[width=\linewidth]{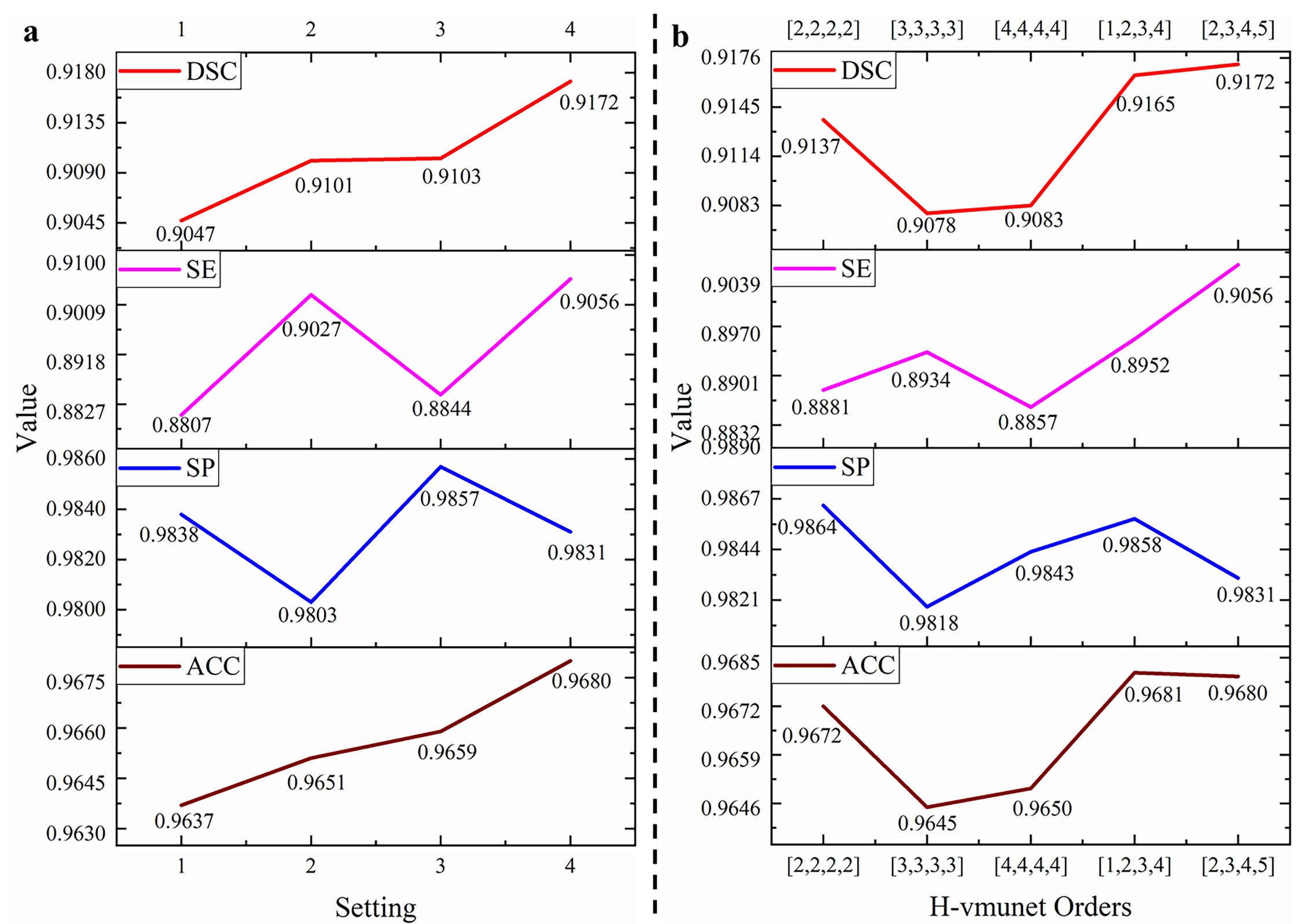}
\caption{(a) Ablation experiments on the effects of employing SS2D operations in High-order spatial interaction mechanisms. (b) Ablation experiments on the effect of SS2D operations of different orders on H-vmunet.}
\label{fig07}
\end{figure}

In order to verify the validity of incorporating the SS2D operation in High-order spatial interaction mechanisms, we conducted ablation experiments. As shown in Figure \ref{fig06}, we as 4 setups respectively, where setup 4 is our proposed H-SS2D module. Setup 3 is to replace the SS2D of each order with a linear projection layer, Setup 2 is to replace the proposed Local-SS2D module with a Global Local Filter \cite{rao2022hornet} module, and Setup 1 is to replace both the SS2D and Local-SS2D modules of each order. The results of our experiments for this ablation are shown in Figure \ref{fig07}(a), where we can conclude that the Dice value is lowest in Setting 1 where neither SS2D is involved. And the best result is reached in setting 4.

In order to verify the effect of different orders of SS2D operation interactions on the H-vmunet, we conducted ablation experiments. The specific experimental setup is shown in Figure \ref{fig07}(b), with H-vmunet Orders of [2,3,4,5] denoting the spatial interactions of the four H-VSS modules in the direction of the input stream of the encoder with orders 2, 3, 4, and 5, respectively, and a symmetric setup for the decoder. From Figure \ref{fig07}(b), the [1,2,3,4] and [2,3,4,5] setups with mixed different orders show better performance than a single identical order interaction. The use of higher orders also introduces more memory consumption, so in this paper we use mixed orders of [2,3,4,5] as the order setting for H-vmunet.

\section{Conclusion}
In this study, we propose a High-order 2D-selective-scan (H-SS2D) based on the state-space model (SSM). The higher-order operation is able to maintain the superior global receptive field of SS2D while minimizing the introduction of redundant information. In addition, H-SS2D is embedded in the proposed higher-order visual state space (H-VSS) module instead of the conventional visual state space (VSS) module. Meanwhile, H-VSS is combined with the UNet framework, and we propose a High-order Vision Mamba UNet (H-vmunet) for medical image segmentation. Currently, we validate the effectiveness of the proposed H-vmunet on three publicly available medical image datasets (ISIC2017, Spleen and CVC-ClinicDB). In addition, the number of parameters of our proposed H-vmunet is reduced by 67.28$\%$ compared to the traditional pure Vision Mamba UNet model (VM-UNet). Since the proposed H-SS2D reduces the introduction of redundant information in the traditional SS2D and enhances the learning of local feature information at the same time, it may become one of the strong contenders of SS2D in the future.

%\bibliographystyle{plain}
%\bibliography{ref}

\end{document}